\documentclass{article}

\usepackage[preprint,nonatbib]{neurips_2025}  
\usepackage{graphicx}
\usepackage{adjustbox}
\usepackage{multirow}
\usepackage{amsmath}
\usepackage[numbers,sort&compress]{natbib}

\usepackage[utf8]{inputenc} 
\usepackage[T1]{fontenc}    
\usepackage{hyperref}       
\usepackage{url}            
\usepackage{booktabs}       
\usepackage{amsfonts}       
\usepackage{nicefrac}       
\usepackage{microtype}      
\usepackage{xcolor}         

\title{OpenRLHF: An Easy-to-use, Scalable\\and High-performance RLHF Framework}

\author{%
  Jian Hu\thanks{Project Leader.}
  \And
  Xibin Wu
  \And
  Wei Shen
  \And
  Jason Klein Liu
  \And
  Zilin Zhu
  \AND
  Weixun Wang
  \And
  Songlin Jiang
  \And
  Haoran Wang
  \And
  Hao Chen
  \And
  Bin Chen
  \AND
  Weikai Fang
  \And
  Xianyu
  \And
  Yu Cao
  \And
  Haotian Xu
  \And
  Yiming Liu
  \AND
  \adjustbox{valign=c,raise=0.4ex}{\includegraphics[scale=0.1]{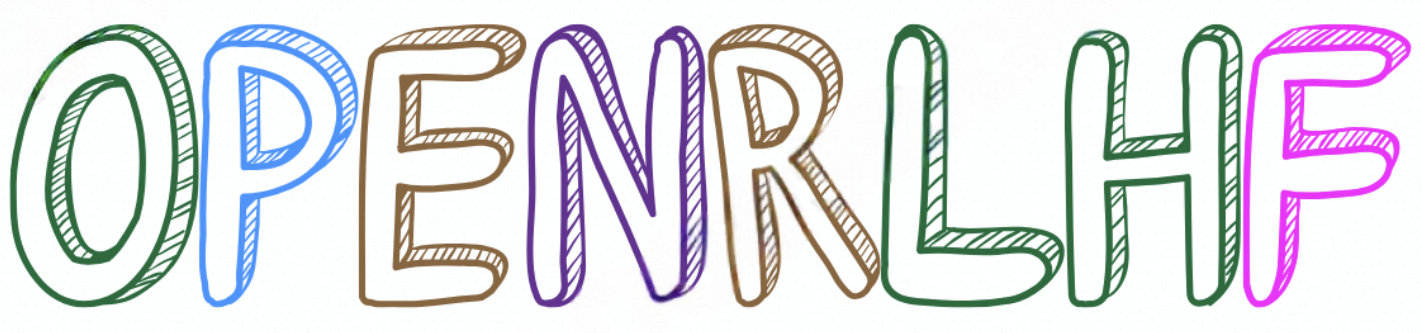}}\quad\large Team\thanks{Full contributor list available in Appendix~\ref{sec:contributors}.}
}

\begin{document}

\maketitle

\begin{abstract}
  Large Language Models (LLMs) fine-tuned via Reinforcement Learning from Human Feedback (RLHF) and Reinforcement Learning with Verifiable Rewards (RLVR) significantly improve the alignment of human-AI values, further raising the upper bound of AI capabilities, particularly in reasoning-intensive, long-context Chain-of-Thought (CoT) tasks. However, existing frameworks commonly face challenges such as inference bottlenecks and complexity barriers, which restrict their accessibility to newcomers. To bridge this gap, we introduce \textbf{OpenRLHF}, a user-friendly, scalable, and easy-to-learn open-source RLHF framework built upon Ray, vLLM, DeepSpeed, and HuggingFace Transformers, featuring a simplified design, clear code structure, and comprehensive documentation to facilitate entry for researchers and practitioners. Experimental results show that OpenRLHF achieves superior training efficiency, with speedups ranging from 1.22× to 1.68× across different model sizes, compared to state-of-the-art frameworks. Additionally, it requires significantly fewer lines of code for implementation. OpenRLHF is publicly available at \url{https://github.com/OpenRLHF/OpenRLHF}, and has already been adopted by leading institutions to accelerate RLHF research and learning.
\end{abstract}

\section{Introduction}

Large Language Models (LLMs) fine-tuned via \textbf{Reinforcement Learning from Human Feedback (RLHF)} and \textbf{Reinforcement Learning with Verifiable Rewards (RLVR)} have markedly advanced human-AI alignment and elevated the upper bound of AI capabilities \cite{christiano2017deep,stiennon2020learning,guo2025deepseek,shen2025exploring}. These approaches enable models to better conform to human intentions and values while achieving superior reasoning performance. Notably, models such as GPT-4~\cite{achiam2023gpt}, DeepSeek-R1~\cite{guo2025deepseek}, and Claude~\cite{askell2021general} excel at complex reasoning tasks by generating detailed step-by-step rationales, commonly referred to as \emph{Chain-of-Thought} (CoT) outputs.

However, RLHF and RLVR training methodologies—especially those employing Proximal Policy Optimization (PPO)—face significant computational challenges. In particular, the inference phase often accounts for over 90\% of the total RLHF (or RLVR) runtime, as models need to generate thousands of tokens during each inference step. Consequently, there is an increasing demand for efficient and scalable frameworks that reduce inference overhead and simplify the training workflows for distributed RLHF and RLVR.

Existing RLHF and RLVR systems, such as DeepSpeed-Chat~\cite{yao2023deepspeed}, TRL~\cite{vonwerra2022trl}, and ColossalChat~\cite{li2023colossal}, have made notable progress in distributed computation and memory efficiency. Conversely, industrial-grade solutions like Nemo-aligner~\cite{shen2024nemoaligner} and ChatLearn~\cite{chatlearn2023} offer advanced optimizations at the modeling and framework levels. While verl~\cite{sheng2024hybridflow}, a framework proposed after our initial development, also provides sophisticated optimizations (e.g., its 3D-Hybrid engine), these industrial solutions generally feature tightly coupled and specialized designs that introduce considerable complexity, steep learning curves, and accessibility barriers for newcomers and academic researchers. Yet, their tightly coupled and specialized designs introduce significant complexity, steep learning curves, and accessibility barriers for newcomers and academic researchers. Therefore, there remains a pressing need for an RLHF and RLVR framework that balances high performance, scalability, and ease of use—one that is straightforward enough for researchers new to the field, yet adaptable to diverse and evolving workloads.

In this paper, we present \textbf{OpenRLHF}, a simple, high-performance, fully open-source framework supporting both RLHF and RLVR, built upon Ray~\cite{liang2018rllib}, vLLM, DeepSpeed~\cite{yao2023deepspeed}, and HuggingFace Transformers~\cite{wolf-etal-2020-transformers}. OpenRLHF offers these key contributions:

\begin{itemize}
    \item \textbf{First Ray-Based Open-Source RLHF and RLVR Architecture}: Leveraging Ray's flexible distributed computing primitives, OpenRLHF enables streamlined orchestration and resource management for RLHF and RLVR workflows, significantly simplifying distributed operations and deployments while enhancing usability and flexibility.
    
    \item \textbf{3D Parallelism with DeepSpeed-ZeRO and Ring Attention}: To enable seamless and efficient scalability for large models, OpenRLHF integrates automatic tensor parallelism (AutoTP) provided by DeepSpeed-ZeRO and implements sequence parallelism via ring attention. This streamlined integration realizes an efficient "3D" parallel strategy—combining tensor, data, and sequence parallelism—without requiring complex engineering or extensive user configuration.
    
    \item \textbf{First Accelerated CoT Inference with vLLM}: Addressing the critical inference bottleneck in long-chain-of-thought RLHF and RLVR workloads, OpenRLHF incorporates the state-of-the-art vLLM inference engine. This integration accelerates inference through token-level parallel decoding, advanced caching, and optimized dynamic batching, thereby substantially improving overall training runtime efficiency.

    \item \textbf{Asynchronous Dataflow and Remote Engine Interactions}: OpenRLHF supports asynchronous dataflow and remote engine communication to maximize system throughput and resource utilization during distributed RLHF training. In this architecture, rollout engines, actor engines, and remote engines operate independently and communicate via message passing, enabling immediate processing as soon as data becomes available. This design reduces idle time, improves pipeline efficiency, and enhances scalability and flexibility across large distributed GPU clusters. By leveraging asynchronous remote engine interactions, OpenRLHF can be easily extended to support scalable \textbf{agent RL} training.
\end{itemize}

Together, these innovations position OpenRLHF as an accessible yet robust framework suitable for both efficient experimentation in academic environments and practical deployment scenarios. Already adopted by leading institutions and companies—including CMU, MIT, Microsoft, and HKUST—OpenRLHF demonstrates broad applicability and impact across the research community (see Appendix~\ref{sec:broad_impact} for details on the framework's broad impact and adoption). We publicly release OpenRLHF to foster openness, accelerate research, and promote innovation within the RLHF and RLVR ecosystem.

\section{Related Work}
\paragraph{RLHF and RLVR} Reinforcement Learning from Human Feedback (RLHF) has emerged as a pivotal paradigm for aligning large language models with human preferences~\cite{christiano2017deep,stiennon2020learning}. The foundational RLHF framework trains a reward model using human preference data and optimizes the language model with reinforcement learning algorithms, such as PPO~\cite{schulman2017proximal}. This approach has been successfully deployed in prominent models, including InstructGPT~\cite{ouyang2022training}, ChatGPT, and GPT-4~\cite{achiam2023gpt}. Building upon RLHF, Reinforcement Learning with Verifiable Rewards (RLVR) leverages automatically verifiable signals from mathematical verification, code execution, or other automated evaluation mechanisms~\cite{guo2025deepseek,shen2025exploring,cobbe2021training}. While PPO remains dominant due to its stability~\cite{schulman2017proximal,ziegler2019fine}, alternative approaches, such as Direct Preference Optimization (DPO)~\cite{rafailov2023direct}, have gained attention for their computational efficiency. However, both RLHF and RLVR methods require substantial computational resources and sophisticated distributed training strategies. The computational challenges have motivated various optimization techniques, including efficient inference engines~\cite{kwon2023efficient}, memory-efficient training strategies~\cite{rajbhandari2020zero}, and distributed orchestration frameworks~\cite{yao2023deepspeed}. Nevertheless, existing solutions often sacrifice either performance for simplicity or accessibility for optimization, creating a gap that OpenRLHF aims to address.

\paragraph{RLHF (RLVR) Frameworks} The computational complexity of RLHF and RLVR training has driven the development of specialized frameworks for these tasks. General-purpose RL frameworks~\cite{liang2018rllib,openai2017baselines} designed for small-scale networks often fail to address LLM-specific challenges and typically employ multi-controller architectures with complex inter-process communication, resulting in steep learning curves. RLHF-specific frameworks face significant performance-accessibility trade-offs. Open-source solutions like TRL~\cite{vonwerra2022trl}, DeepSpeed-Chat~\cite{yao2023deepspeed}, and ColossalChat~\cite{li2023colossal} provide accessible implementations but often lack sophisticated orchestration capabilities and struggle with inference optimization. Industrial frameworks like Nemo-aligner~\cite{shen2024nemoaligner}, ChatLearn~\cite{chatlearn2023}, and verl~\cite{sheng2024hybridflow} offer advanced optimizations including 3D parallelism and memory management. However, they feature tightly coupled architectures requiring substantial engineering expertise and extensive infrastructure setup, creating accessibility barriers for academic researchers. These systems employ static resource allocation paradigms, resulting in suboptimal utilization and limited adaptability. Most frameworks inadequately address both inference bottlenecks and usability challenges simultaneously, forcing users to choose between high performance and ease of use.

\section{Design of OpenRLHF}

\subsection{Overview: Ray-based RLHF architecture}
\begin{figure}[!htbp]
    \centering
    \includegraphics[width=0.6\linewidth]{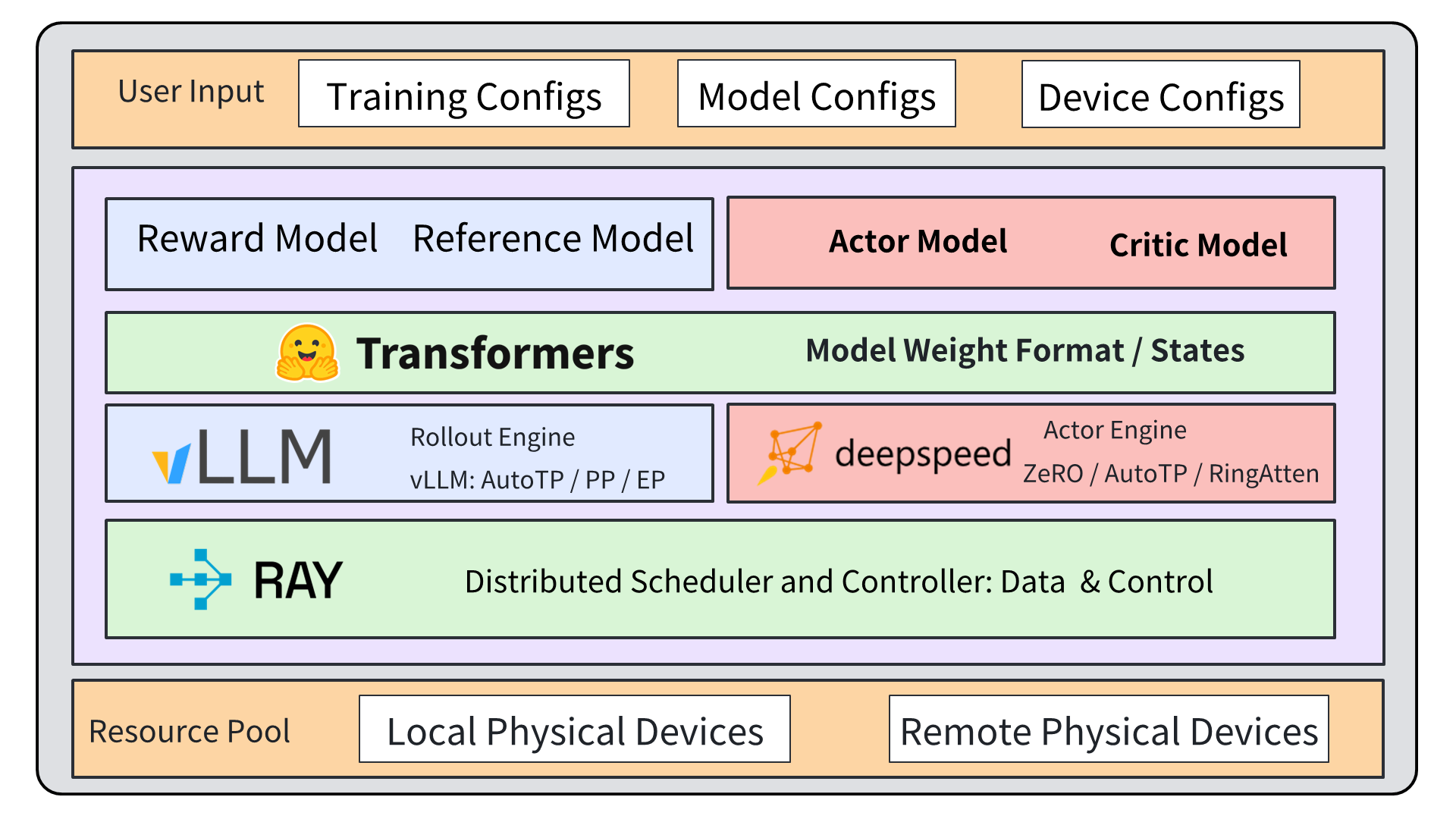}
    \caption{Overall architecture of OpenRLHF. The system assigns GPUs to two primary roles: rollout engines dedicated to response generation and actor engines responsible for computing log-probabilities and model training. OpenRLHF leverages Ray for distributed scheduling, integrates vLLM to achieve efficient response rollout with low GPU memory usage, and employs DeepSpeed ZeRO for 3D parallelism (including tensor, data, and sequence parallelism) to enable efficient training. The underlying models are instantiated with flexible Transformer architectures, making the system easy to extend and adapt for diverse scenarios.}
    \label{fig:openrlhf_design}
\end{figure}
OpenRLHF is the first open-source, Ray-based RLHF architecture that assigns a batch of GPUs to distinct roles and manages both data flow and workflow among these roles using Ray's scheduling capabilities. As illustrated in Figure~\ref{fig:openrlhf_design}, it defines two primary roles: the \textbf{rollout engine}, responsible for response generation to given prompts, and the \textbf{ZeRO engine}, which computes \textit{logprobs}, \textit{reference policy logprobs}, and handles model training (For a detailed design of the PPO workflow, refer to Appendix~\ref{sec:ppo_workflow}, and for in-depth implementation tips, consult Blog \cite{shen2024advanced}.)

Additionally, we leverage \textbf{vLLM} as the rollout engine, enabling efficient response generation with minimal GPU memory usage. For model training, we adopt DeepSpeed, which implements 3D parallelism, including automatic tensor parallelism, ZeRO/data parallelism, and sequence parallelism, to train both the actor and value models efficiently.

\paragraph{Core reason for ease of use} The remarkable usability of our framework stems from three core design principles: simplified model slicing, seamless integration, and flexible scheduling, which collectively streamline the workflow and significantly reduce the implementation burden during training and deployment. The exchange of model weights between the rollout engine and the training engine is enabled by a flexible slicing and partitioning pipeline. Hugging Face Transformer~\cite{jain2022hugging} models are instantiated and trained using \textbf{DeepSpeed ZeRO}, \textbf{AutoTP}, and \textbf{Ring-Attention}~\cite{liu2023ring} model parallelism. These model slices are then efficiently transferred to vLLM through AutoTP and AutoPP, which dynamically partition the models into sub-modules. The \textbf{Ray-based scheduling mechanism} enables seamless switching between different model parallelism modes, such as hybrid engine and asynchronous training. This streamlined workflow significantly reduces complexity, making the system highly user-friendly and easy to extend. Compared with architectures such as DeepSpeed-Chat, Transformer Reinforcement Learning (TRL), or other mainstream frameworks, OpenRLHF supports asynchronous dataflow and remote engine interactions, significantly improving the overall efficiency of the training process and the agent workflow.

\subsection{Distributed and Efficient System Design}

\paragraph{3D Parallelism with DeepSpeed ZeRO and Ring Attention} To enable seamless and efficient scalability for large models, OpenRLHF integrates the latest automatic tensor parallelism (AutoTP) feature from DeepSpeed ZeRO. In many industrial-grade RLHF architectures, users previously needed to manually specify an injection policy for each transformer model, identifying the linear layers and attention outputs that required communication between data-parallel ranks. In contrast, OpenRLHF leverages DeepSpeed ZeRO's new capability to support automatic tensor parallelism for HuggingFace models by default. When kernel injection is not enabled and an injection policy is not provided, DeepSpeed automatically determines and applies the necessary policy at runtime. It can dramatically simplify the user experience and extend robust tensor parallelism support to a broader range of models, removing the need for complex engineering or manual configuration.

In addition, OpenRLHF implements sequence parallelism through ring attention. Ring attention employs a ring-based communication topology, efficiently distributing attention computation for long sequences across multiple GPUs while minimizing both memory usage and communication overhead. It is especially critical for modern RLHF and RLVR workloads involving long CoT reasoning, where conventional attention computation can become a major scalability bottleneck. By combining AutoTP, data parallelism, and ring attention-based sequence parallelism, OpenRLHF empowers large-scale, efficient, and highly usable RLHF model training on flexible GPU clusters.

\paragraph{Accelerated CoT Inference with vLLM} As LLMs advance in reasoning, RLHF and RLVR pipelines increasingly face inference bottlenecks, particularly with long CoT outputs. For models like OpenAI-o1 and DeepSeek-R1, CoT generation can dominate training time, making efficient long-form inference crucial for scalability. To address this challenge, OpenRLHF integrates the vLLM inference framework, which is specifically designed for high-throughput and memory-efficient LLM serving. vLLM provides a streamlined interface for generating RLHF samples and supporting frequent model weight updates.

The core innovation in vLLM is efficient management of attention key and value memory with \textbf{PagedAttention}~\cite{kwon2023efficient}. The technique significantly reduces memory waste to less than 4\%, enabling the batching of more sequences and enhancing GPU utilization and throughput. PagedAttention also supports efficient memory sharing for advanced sampling methods, such as parallel sampling and beam search, reducing memory usage by up to 55\% and further boosting inference efficiency. Besides PagedAttention, vLLM has several other advantages, including continuous batching of incoming requests, fast model execution with CUDA Graph, and CUDA kernels optimized with FlashAttention and FlashInference, which enable rapid and memory-efficient attention computations. It also features speculative decoding for faster inference and chunked prefill to reduce latency on long sequences. Collectively, these enhancements make vLLM highly effective for large-scale, long-sequence inference, especially suitable for RLHF and RLVR pipelines, where efficiency and scalability are crucial.

\paragraph{Asynchronous Dataflow and Remote Engine Interactions} OpenRLHF supports asynchronous dataflow and remote engine communication to maximize system throughput and resource utilization during distributed RLHF training. In this architecture, rollout engines, actor engines, and remote engines operate independently and communicate via message passing, enabling immediate processing as soon as data becomes available. 
Fully asynchronous execution is critical in the CoT era, where inference involves generating multi-step reasoning that can vary significantly in length and computational cost. In synchronous frameworks, the slowest CoT generation can bottleneck the whole pipeline and waste resources. In contrast, OpenRLHF's asynchronous design allows each engine to operate at its own pace, ensuring hardware is utilized even for long or variable CoT tasks. It not only accelerates training and evaluation but also enables efficient scaling and supports dynamic agent RL workflows. Leveraging asynchronous remote engine interactions, OpenRLHF is readily extensible for scalable \textbf{agent RL} training in modern, CoT-centric environments.

\begin{table*}[!ht]
\centering
\footnotesize
\setlength{\tabcolsep}{6pt}
\begin{tabular}{l|cc|cc|cc|cc|c}
\toprule
\multirow{3}{*}{\textbf{Model Size}} & \multicolumn{2}{c|}{\textbf{1K}} & \multicolumn{2}{c|}{\textbf{2K}} & \multicolumn{2}{c|}{\textbf{4K}} & \multicolumn{2}{c|}{\textbf{8K}} & \multirow{3}{*}{\textbf{Avg. Speedup}} \\
& \textbf{Ours} & \textbf{verl} & \textbf{Ours} & \textbf{verl} & \textbf{Ours} & \textbf{verl} & \textbf{Ours} & \textbf{verl} & \\
& \textbf{(sec)} & \textbf{(sec)} & \textbf{(sec)} & \textbf{(sec)} & \textbf{(sec)} & \textbf{(sec)} & \textbf{(sec)} & \textbf{(sec)} & \\
\midrule
1.5B  & 14.9 & 16.2 & 26.5 & 33.1 & 61.6  & 65.5  & 113.0 & 134.8 & 1.22$\times$ \\
7B    & 16.0 & 17.3 & 30.3 & 47.3 & 90.3  & 101.3 & 226.4 & 232.4 & 1.42$\times$ \\
14B   & 25.5 & 28.5 & 51.0 & 74.3 & 136.3 & 202.8 & 328.6 & 511.1 & 1.68$\times$ \\
\bottomrule
\end{tabular}
\caption{The average training time (seconds) per step for different model sizes and context window lengths compared between OpenRLHF and verl. The speedup is calculated as the geometric mean of verl time / OpenRLHF time across all sequence lengths.}
\label{table:verl_performance}
\end{table*}

\section{Experiments}

\subsection{Performance Comparison}\label{sec:cot}

\paragraph{Long CoT Experiment Setup} To ensure the applicability of the compared methods in current RLHF workflows, we conduct our experimental evaluation under a long-chain-of-thought (CoT) generation scenario. Given computational resource constraints, we focus our comparison on the long CoT RLVR setting, benchmarking OpenRLHF against verl, currently the state-of-the-art framework for RLHF training. We evaluate the training efficiency of OpenRLHF (v0.8.5) and verl (v0.4.0) by measuring the average per-step training time (in seconds) across various model sizes (1.5B, 7B, and 14B parameters) and maximum generation lengths (1K, 2K, 4K, and 8K tokens). To ensure the base models can produce sufficiently long contextual outputs for stress testing, we adopt the DeepSeek open-source distilled Qwen series. All models are fine-tuned using the Decoupled Clip and Dynamic Sampling Policy Optimization (DAPO) algorithm \cite{yu2025dapo} under identical hyperparameter settings. Experiments are conducted on 8 NVIDIA H200 140GB GPUs using PyTorch 2.7 \cite{imambi2021pytorch} and the ZeRO Stage 3 optimizer or Fully Sharded Data Parallel (FSDP) \cite{zhao2023pytorch}. For each configuration, the local batch size is set to 1 to mitigate the risk of out-of-memory errors, with a maximum input context length of 1024 tokens. Following the discussion in \cite{liu2025rethinking}, we use $k_{2}$ as the loss function. The reported values represent the average training time per step, excluding the first 10 steps.

\paragraph{Long CoT Performance Analysis} The experimental results in Table~\ref{table:verl_performance} show that OpenRLHF consistently achieves superior training speed compared to verl across all configurations. OpenRLHF delivers speedups ranging from 1.22× for the 1.5B model to 1.68× for the 14B model, with performance advantages becoming more pronounced as model size and context length increase. For instance, in the 14B-8K setting, OpenRLHF achieves a 1.56× speedup (328.6 seconds vs. 511.1 seconds), while the 7B-2K configuration shows a 1.56× speedup (30.3 seconds vs. 47.3 seconds). These speedup improvements can be attributed to OpenRLHF's algorithmic design, including the DAPO optimization strategy, which effectively mitigates memory overhead and computational bottlenecks under long-context scenarios. The consistent speedup across different scales underscores OpenRLHF's efficiency advantages in contemporary RLHF pipelines.



\paragraph{General RLVR Experiment} To ensure a fair and controlled evaluation of training efficiency in reinforcement learning fine-tuning, we compare OpenRLHF with the optimized TRL framework on the GSM8K dataset \cite{cobbe2021training} using the GRPO algorithm over a single training epoch. Both systems are configured with identical hyperparameters and run on the same hardware setup to isolate the effect of framework design on performance. The GRPO algorithm is selected due to its relevance in reward-based fine-tuning scenarios, and GSM8K serves as a representative benchmark for arithmetic reasoning tasks. In terms of training efficiency, OpenRLHF demonstrates a substantial advantage, completing one epoch in 1,657 seconds, compared to the 5,189 seconds required by TRL, representing approximately a 3.1× speedup. The result highlights the superior efficiency and maintainability of OpenRLHF's implementation, which benefits from a streamlined design and targeted system-level optimizations.

\paragraph{General RLHF Experiment} To evaluate the training efficiency of modern RLHF frameworks, we compare OpenRLHF with the optimized DeepSpeed-Chat (DSChat) implementation. The experiment involves fine-tuning 1,024 prompts using the PPO algorithm for one epoch under identical hardware and hyperparameter configurations. This setup ensures that observed differences in performance are attributable solely to differences in system design and optimization strategies between the two frameworks.
In terms of training time, OpenRLHF completes the task in 236.8 seconds, significantly outperforming DSChat, which requires 855.09 seconds, resulting in a 3.6× speedup. This performance gain is primarily driven by two system-level innovations in OpenRLHF: the use of vLLM for accelerated token generation and Ray for efficient distributed execution. Collectively, these design choices result in a more scalable and excellent RLHF framework.

\subsection{Usability Comparison}

\begin{figure}
    \centering
    \includegraphics[width=0.6\linewidth]{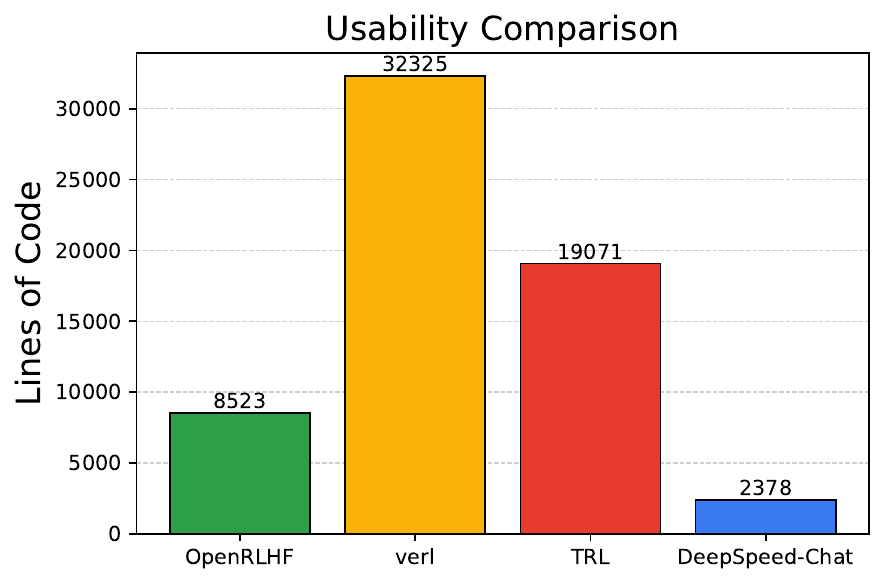}
    \caption{Core code complexity comparison across RLHF frameworks (lines of code). Lower values indicate simpler implementation and better maintainability.}
    \label{fig:usability_comparison}
\end{figure}

As illustrated in Figure~\ref{fig:usability_comparison}, OpenRLHF achieves a compelling balance between implementation conciseness and training performance. Despite being the second most concise framework with only 8,523 lines of code—significantly fewer than TRL (19,071) and verl (32,325)—OpenRLHF demonstrates a clear performance advantage across standard RLHF benchmarks. This streamlined codebase not only facilitates easier comprehension and modification for developers but also reduces the engineering overhead associated with integration into custom pipelines. In addition to its lightweight design, OpenRLHF offers comprehensive support for various reinforcement learning fine-tuning paradigms, including Supervised Fine-Tuning~(SFT), Direct Preference Optimization~(DPO) \cite{rafailov2023direct}, Reward Model~(RM), and Process Reward Model~(PRM) \cite{lightman2023let}. This broad functionality, combined with its modular and well-documented architecture, significantly lowers the barrier to entry for both research and production use. Overall, OpenRLHF's design exemplifies high usability by combining minimal code complexity with extensive functionality and competitive performance.

\section{Limitations}
While OpenRLHF demonstrates significant advantages in balancing performance and accessibility, several limitations should be acknowledged. Despite our optimization efforts, OpenRLHF may not match the peak performance of highly specialized industrial frameworks that benefit from dedicated engineering teams and extensive resources. As a community-driven, open-source project without dedicated economic support, OpenRLHF faces resource constraints in rapidly integrating cutting-edge features, which may result in delays compared to well-funded commercial frameworks. Currently, the framework primarily focuses on language models and does not support Vision-Language Models or other multimodal architectures, thereby limiting its applicability to multimodal AI alignment research. Additionally, OpenRLHF's modular design introduces dependencies on external systems such as Ray, vLLM, and DeepSpeed, where updates in these upstream systems may require maintenance work or introduce compatibility issues. Despite these limitations, we believe OpenRLHF's contributions to accessibility and democratization of RLHF research provide significant value to the community.

\section{Conclusion}
We presented OpenRLHF, a simple yet high-performance open-source framework that bridges the gap between performance and usability in RLHF and RLVR training. By integrating Ray's distributed computing, vLLM's inference optimization, DeepSpeed ZeRO's memory efficiency, and ring attention's sequence parallelism, OpenRLHF delivers four key innovations: a Ray-based architecture for streamlined orchestration, 3D parallelism for efficient scaling, accelerated CoT inference that addresses the critical bottleneck, and asynchronous dataflow for maximum throughput. The framework's broad adoption across leading institutions and companies—from academic courses at CMU to production deployments at major tech companies—validates its real-world effectiveness. OpenRLHF's influence on subsequent frameworks and its role in democratizing RLHF research demonstrate its significant contribution to the field. By open-sourcing this framework, we aim to accelerate research progress and enable broader participation in the development of aligned AI.

\clearpage

\bibliographystyle{unsrtnat}
\bibliography{reference}

\clearpage
\appendix

\section{Full Contributors}\label{sec:contributors}

A more complete list can be found in the OpenRLHF commit and release history.

\paragraph{Ray Integration} Jian Hu, Xibin Wu, Songlin Jiang
\paragraph{vLLM Integration} Jian Hu, Xibin Wu, Songlin Jiang, Kaichao You (vLLM Team)
\paragraph{Ring Attention} Zilin Zhu (Zhipu), Zhibo Zhou (Vivo), gzpan(GitHub User), Jian Hu 
\paragraph{Supervised Learning:} Haoran Wang and Xianyu
\paragraph{DeepSpeed Integration} Jian Hu, Xibin Wu, Songlin Jiang, Bin Chen, Yiming Liu
\paragraph{Asynchronous Agentic RL} Haotian Xu and Jian Hu
\paragraph{Single Controller Architecture} Jian Hu
\paragraph{PPO Implementation} Jian Hu
\paragraph{GRPO Implementation} Jason Klein Liu
\paragraph{KL Control Mechanism} Yiming Liu and Jason Klein Liu
\paragraph{RLVR Experiment Design} Jason Klein Liu
\paragraph{RLHF Experiment Design} Jian Hu
\paragraph{Dynamic Sampling} Jian Hu
\paragraph{Dynamic Batching} Hao Chen
\paragraph{Documentation} Jian Hu
\paragraph{Testing and Bug Reports} Weikai Fang, Wei Shen, Weixun Wang
\paragraph{Paper Writing and Presentations} Wei Shen, Jason Klein Liu, Weixun Wang, Yu Cao, Jian Hu

\section{Broad Impact and Adoption}\label{sec:broad_impact}

OpenRLHF has achieved significant adoption and impact across both academic and industrial communities since its release. The framework's balance of high performance and accessibility has made it a preferred choice for diverse applications ranging from research experimentation to production deployment.

OpenRLHF has been formally integrated into academic curricula, notably being adopted by the CMU Advanced Natural Language Processing course in Spring 2025 as a core teaching framework. This integration demonstrates the framework's educational value and accessibility for students new to RLHF. The framework has also garnered recognition from major technology communities, including an invitation to present at the PyTorch Expert Exchange 2025, which highlights its technical contributions to the broader deep learning ecosystem.

The framework has been widely adopted by leading technology companies and research institutions, including Google, ByteDance, Baidu, NVIDIA, Tencent, China Telecom, Vivo, NexusFlow, JSC, UC Berkeley's Starling Team, Meituan, and HKUST. This diverse adoption across major tech companies, telecommunications providers, and academic institutions demonstrates OpenRLHF's versatility and robustness in handling various scales of RLHF workloads, from research prototyping to production-scale deployments.

OpenRLHF's modular design and extensible architecture have enabled it to serve as a foundation for specialized frameworks. Notable examples include LMM-R1 for multimodal reinforcement learning, MARTI for advanced reasoning tasks, and MM-EUREKA for multimodal applications. These derivative frameworks demonstrate OpenRLHF's flexibility in supporting diverse research directions and its role as a platform for innovation in the RLHF ecosystem.

The design principles and technical innovations introduced in OpenRLHF have influenced subsequent framework development in the field. Several prominent frameworks, including verl, Alibaba ROLL, SLIME, and Open-Reasoner-Zero, have acknowledged OpenRLHF's contributions in their documentation and publications, citing its distributed architecture design, Ray-based orchestration approach, and integration strategies as influential to their own development. This acknowledgment reflects OpenRLHF's role in advancing the state-of-the-art in RLHF system design and establishing best practices for distributed RLHF training.

Beyond direct usage and citations, OpenRLHF has contributed to democratizing RLHF research by lowering the barrier to entry for academic researchers and smaller organizations. The framework's emphasis on ease of use without sacrificing performance has enabled broader participation in RLHF research, fostering innovation across diverse research communities that might otherwise lack the resources for complex distributed training setups.

\begin{figure*}[!htbp]
    \centering
    \includegraphics[width=0.8\linewidth]{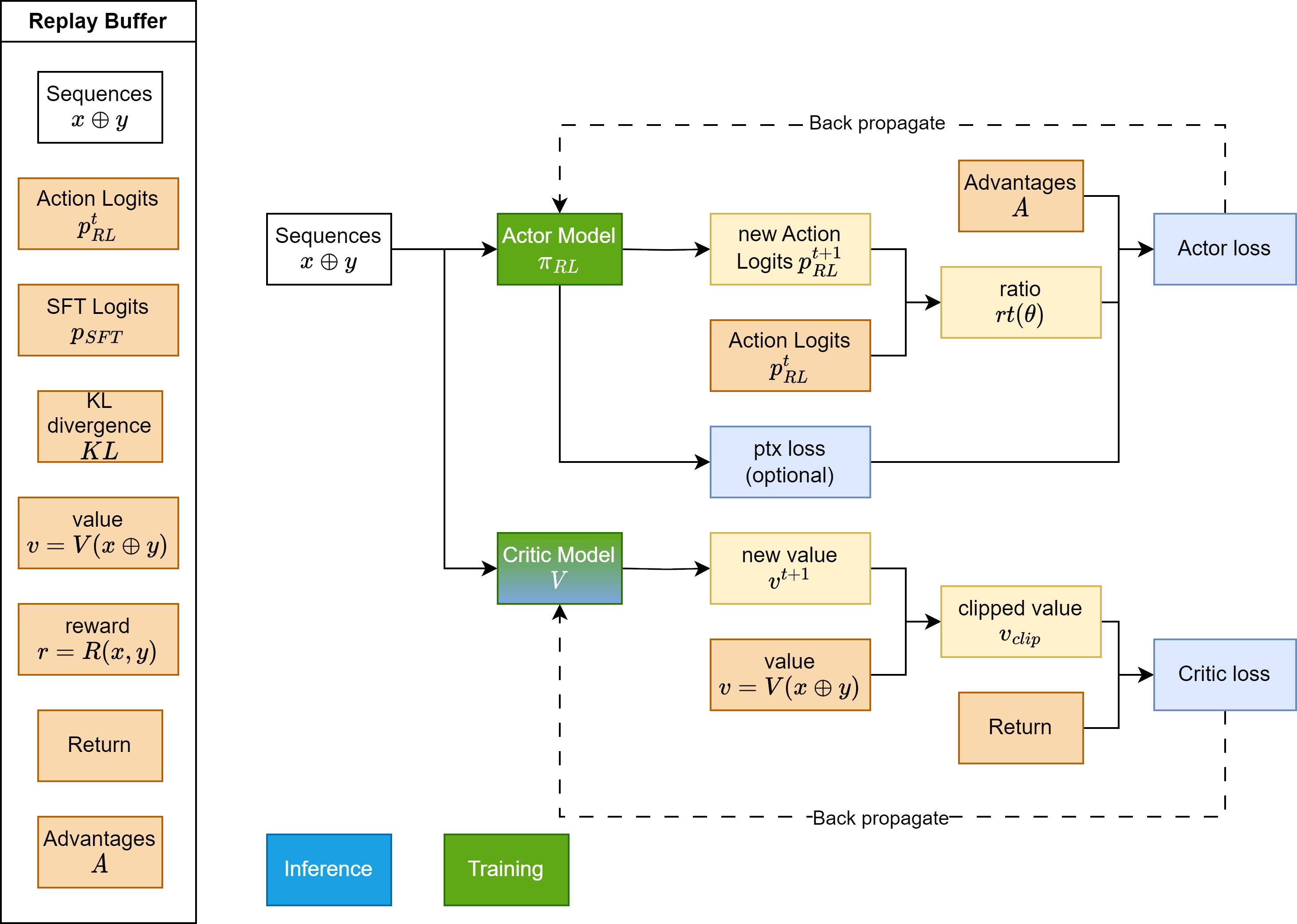}
    \caption{Overall PPO workflow of OpenRLHF.}
    \label{fig:openrlhf_arc}
\end{figure*}

\section{PPO Workflow Design}\label{sec:ppo_workflow}

This section presents OpenRLHF's comprehensive PPO-based RLHF training workflow, which orchestrates multiple specialized engines to efficiently handle the complex multi-stage training process (as shown in Figure \ref{fig:openrlhf_arc}). The OpenRLHF PPO workflow consists of four main stages executed iteratively: (1) \textbf{Rollout Generation}, where the current policy generates responses to prompts; (2) \textbf{Reward Computation}, where responses are evaluated using a trained reward model; (3) \textbf{Advantage Estimation}, where advantages and returns are calculated using GAE (Generalized Advantage Estimation); and (4) \textbf{Policy Optimization}, where the policy is updated using PPO loss. This pipeline repeats for multiple iterations until convergence is achieved.

The training begins with the \textbf{Rollout Engine} generating responses for a batch of prompts using the current policy $\pi_\theta$. A batch of prompts $\{x_1, x_2, ..., x_B\}$ is sampled from the training dataset, and the Rollout Engine, equipped with vLLM for efficient inference, generates responses $\{y_1, y_2, ..., y_B\}$ using the current policy. During generation, the engine records action log-probabilities $\log \pi_\theta(y_i|x_i)$ and attention masks. The generated sequences $(x_i, y_i)$ along with their metadata are collected for subsequent processing. The Rollout Engine leverages vLLM's optimizations, including continuous batching, KV-cache management, and PagedAttention, to maximize throughput during this inference-heavy stage.

Once rollouts are generated, the system computes rewards and reference policy log-probabilities. The trained reward model $R_\phi$ evaluates each prompt-response pair to produce scalar rewards $r_i = R_\phi(x_i, y_i)$. Simultaneously, the frozen reference policy $\pi_{\text{ref}}$ computes log-probabilities $\log \pi_{\text{ref}}(y_i|x_i)$ for KL regularization, and the critic network $V_\psi$ estimates state values $V_\psi(x_i, y_{i,:t})$ for advantage computation. These computations can be parallelized across multiple GPUs and are often batched together for efficiency.

The system then computes advantages using Generalized Advantage Estimation (GAE). First, temporal difference residuals are calculated: $\delta_t = r_t + \gamma V_\psi(s_{t+1}) - V_\psi(s_t)$. Then, advantages are computed using GAE: $A_t = \sum_{l=0}^{\infty} (\gamma \lambda)^l \delta_{t+l}$, and discounted returns are calculated as $R_t = A_t + V_\psi(s_t)$. The advantage computation incorporates KL penalty terms to prevent the policy from deviating too far from the reference policy: $r'_t = r_t - \beta \text{KL}[\pi_\theta || \pi_{\text{ref}}]$.

The final stage updates the policy using PPO's clipped objective function. The \textbf{ZeRO Engine} computes the policy loss using the clipped surrogate objective: $L^{\text{CLIP}}(\theta) = \mathbb{E}[\min(r_t(\theta)A_t, \text{clip}(r_t(\theta), 1-\epsilon, 1+\epsilon)A_t)]$, where $r_t(\theta) = \frac{\pi_\theta(a_t|s_t)}{\pi_{\theta_{\text{old}}}(a_t|s_t)}$ is the probability ratio. The critic loss is computed as $L^{V}(\psi) = \mathbb{E}[(V_\psi(s_t) - R_t)^2]$. The critic loss is computed as $L^{V}(\psi) = \mathbb{E}[(V_\psi(s_t) - R_t)^2]$. The total loss combines policy and value losses: $L_{\text{total}} = L^{\text{CLIP}} + c_1 L^{V} + c_2 S[\pi_\theta](s_t)$, where $c_1$ and $c_2$ are coefficients for value loss and entropy bonus respectively, and $S[\pi_\theta](s_t)$ is the entropy of the policy distribution to encourage exploration. The ZeRO Engine performs gradient computation and model parameter updates using DeepSpeed ZeRO optimizations for memory efficiency and scalability.

Throughout this workflow, OpenRLHF's Ray-based architecture enables seamless coordination between different engines while maintaining computational efficiency. The Rollout Engine and ZeRO Engine can operate on different GPU clusters, each optimized for their respective workloads: inference-optimized hardware for rollout generation and training-optimized hardware for policy updates. Ray's distributed scheduling automatically handles data transfer, synchronization, and fault tolerance across the distributed system. This asynchronous execution model enables the system to overlap computation stages whenever possible, thereby significantly improving overall training throughput compared to traditional synchronous RLHF implementations.

The entire PPO training loop continues for multiple epochs, with each epoch processing multiple batches of rollout data. Early stopping criteria based on KL divergence thresholds and performance metrics prevent policy collapse and ensure stable training. The modular design enables the easy integration of advanced techniques, such as advantage normalization, gradient clipping, and adaptive KL penalty coefficients, making OpenRLHF highly customizable for various research and production scenarios.

\end{document}